# Backpropagation Training for Fisher Vectors within Neural Networks

Patrick Wieschollek*  Fabian Groh*  Hendrik P.A. Lensch

University of Tübingen

## Abstract

*Fisher-Vectors (FV) encode higher-order statistics of a set of multiple local descriptors like SIFT features. They already show good performance in combination with shallow learning architectures on visual recognitions tasks. Current methods using FV as a feature descriptor in deep architectures assume that all original input features are static. We propose a framework to jointly learn the representation of original features, FV parameters and parameters of the classifier in the style of traditional neural networks. Our proof of concept implementation improves the performance of FV on the Pascal Voc 2007 challenge in a multi-GPU setting in comparison to a default SVM setting. We demonstrate that FV can be embedded into neural networks at arbitrary positions, allowing end-to-end training with backpropagation.*

## 1. Introduction

Many fundamental computer-vision problems rely on extracting multiple meaningful local *rigid* descriptors like SIFT [12], GIST [13] or SURF [1] features from a single RGB image. Classifying a combination of these features with methods such as Multi-Layer-Perceptron (MPL) or linear Support-Vector-Machines (SVM) often result in good performance [3].

Local features as a compressed image representation are an important factor in visual recognition tasks. Methods like deep *neural networks* (NN) are widely used to *learn* good feature representations without hand-engineering effort. Applied to datasets of large amounts of labeled examples they define state-of-the art results in various computer vision tasks and even surpass human performance [4].

Enriched features, which additionally encode higher-order statistics of the underlying distribution, further improve the classification results. Prominent examples are VLAD [9] and Fisher-Vectors (FV) [14], where the latter is based on a Gaussian-Mixture-Model (GMM) fitted to the data and encodes information relative to each Gaussian component.

The combination of (convolutional) NN and FV using shallow architectures recently became popular [24, 11, 5, 3, 2]. In a nutshell, these methods approach a minimization of the empirical risk $R_{\text{emp}}$, depending on static input features $x \in \mathbb{R}^D$, a feature transformation $f_W(\cdot)$ with parameters $W \in \mathbb{R}^{D \times D}$, a kernel $F$ with parameter $\xi \in \mathbb{R}^{D'}$ and classifier parameters $\theta \in \mathbb{R}^{D'}$ solving

$$(W^\star, \xi^\star, \theta^\star) := \arg\min_{W, \xi, \theta} R_{\text{emp}}(F_\xi(f_W(x)), \theta). \quad (1)$$

All previous methods perform a greedy-wise optimization of these parameters one after another. Usually these steps are: Training neural network parameters $\hat{W}$ for feature extraction on some loss functions, learning GMM components $\hat{\xi}$ on these fixed features $f_{\hat{W}}(x)$ using *expectation-maximization* and finally optimizing a linear SVM to obtain $\hat{\theta}$. Empirical results of this sequential approach already improves performance compared NN. However, it seems reasonable to share information between these optimization steps in the fashion of deep neural networks to *jointly* solve problem (1) in $W, \xi, \theta$.

Therefore, we propose a neural-network-like batch-wised back-propagation training for optimizing $W, \xi, \theta$ together, with a strong theoretical support of [23]. Unfortunately, directly tackling Eq. (1) in a joint optimization approach comes at the price of handling a huge amount of input data. A single image is described by $T$ SIFT features, $T > 8 \cdot 10^4$. For 5k Images from the Pascal Voc 2007 data set this results in approx. 204GB, in contrast to a single 4128-dimensional FV per image. Fortunately, exploiting multi-GPU and sampling approaches makes it possible to compute all necessary parameter update rules in reasonable time.

Our main contributions in this paper can be summarized as follows:

– The proposed architecture includes FV, GMM, and normalization as neural network modules, which enables end-to-end learning in the fashion of classical neural networks.
– We provide the first multi-GPU accelerated implementation for FV computation in large-scale classification tasks.

---

*Indicates equal contribution.

- We introduce feature learning from FV classification including all back-propagation rules to update feature representation.
- The proposed method includes a re-formulation of supervised GMM parameters learning which satisfies GMM constraints naturally using batches.

The remainder of this paper is organized as follows: After delimiting our work from related methods in this field in Section 2, we describe the Fisher-Vector encoding that is used for learning features in Section 3. Section 4 contains details for the back-propagation pass and the learning environment. We evaluate the proposed method in Section 5 and provide concluding remarks in the last Section 6.

## 2. Related Work

The interest of re-using activation values of NN-layers as mid-level features for training additional classifiers in combination with Fisher-Vector and VLAD became popular in recent work. Even for large-scale problems, methods like sparse Fisher-Vector-Coding [11] yield state-of-the-art results in generic object recognition and scene classification [5]. Thereby, computing a FV is usually done as an independent step decoupled from the feature learning process. A kind of stacking of Fisher-Vector layer in deep neural networks was applied to the ILSVRC-2010 challenge with impressive results in [20]. Still, their training method is a greedy layer-by-layer training without back-propagation training steps, which disconnects FV from the already trained layers in the network. Another work considering Fisher-Vectors as a pre-processing step followed by dimensionality reduction methods and a multi-layer-perceptron was recently discussed in [16], again without back propagating updates. In addition, there is no reason for extracted mid-level features from a neural network being the best choice in a completely different classification method. The goal of sharing gradient information between the classifier and Fisher-Vector parameters was first addressed in [21] by adapting the underlying GMM parameters from the classifier loss information. Their algorithm samples GMM parameter gradients from all inputs. To guarantee a decreasing loss they determine the optimal update by line-search. From the computational aspect their approach is limited to learn the Fisher-Kernel only due to the enormous amount of gradient data, which has to be calculated.

Armed with the knowledge of the classifier loss, one might ask how the points of the data set should move to allow the classifier to better separate classes. Common approaches do not incorporate this information shared between the feature learning stage of original feature representation and the classifier. Although it is not possible to shift data points directly, this mapping can be approximated by training $f_W(\cdot)$ in the feature learning stage. This gradient propagation backwards through the FV layer closes the gap between the current *one-direction* usage of FV and deep neural networks. However, any back-propagation through the FV layer ends up with a non-trivial mapping into more than half a million elements. Currently, we are only aware of batch-wise methods to tackle this issue.

Using current methods comes along with several additional drawbacks like the requirement of multiple independent pipelines and limiting the solution space when optimizing (1) and discarding any relevant information like $\nabla \theta$ from the classifier.

Applying these methods to normalized inputs falls into the class of learning methods, where a SVM seeks for the optimal separation hyperplane reducing the theoretical upper-bound for the expected risk $R_{\text{ex}}$, which depends on the radius of the sphere containing all data points (see Theorem 2.1 in [23]). Our work builds on previous attempts without violating this theorem and, thereby, obtaining a strong theoretical basis.

Overall, it is preferable to fully embed FV *within* deep architectures to enable deep *end-to-end* learning approaches without decoupled stages as illustrated in Figure 1 for optimizing the complete set of parameters simultaneously.

## 3. Background

Before deriving the update rules for the Fisher-Vector layer, we shortly recap the analysis and motivation of Fisher-Vectors to be self contained. A comprehensive essay and more details can be found in [22].

**Fisher Vectors** Let $\mathcal{X}$ be a set of features

$$\mathcal{X} = \{x_1, x_2, \dots, x_T\}, \quad x_j \in \mathbb{R}^D$$

for a single image. In computer vision tasks $\mathcal{X}$ typically represents local image descriptors, e.g., 128-dimensional SIFT features. We denote the probability density function as $c_\lambda(\cdot)$ which corresponds to the observations $\mathcal{X}$. The observation $\mathcal{X}$ implies a score function

$$G_\lambda^\mathcal{X} = \nabla_\lambda \ln c_\lambda(\mathcal{X}),$$

with dimension only depending on the number of parameters, not the size of the sample. Let $F_\lambda$ be the Fisher information matrix of $c_\lambda$, a natural kernel (see [8]) for measuring the similarity between two samples $X, Y$ is given by

$$K(X,Y) = \left(G_\lambda^X\right)^T F_\lambda^{-1} G_\lambda^Y,$$
$$F_\lambda = \mathbb{E}_{x \sim c_\lambda} \left[\nabla_\lambda \ln c_\lambda(x) \left(\nabla_\lambda \ln c_\lambda(x)\right)^T\right].$$

Learning a classifier with kernel $K(X, Y)$ is equivalent to learning a linear classifier on the Fisher Vectors $\mathcal{F}_\lambda^X =$

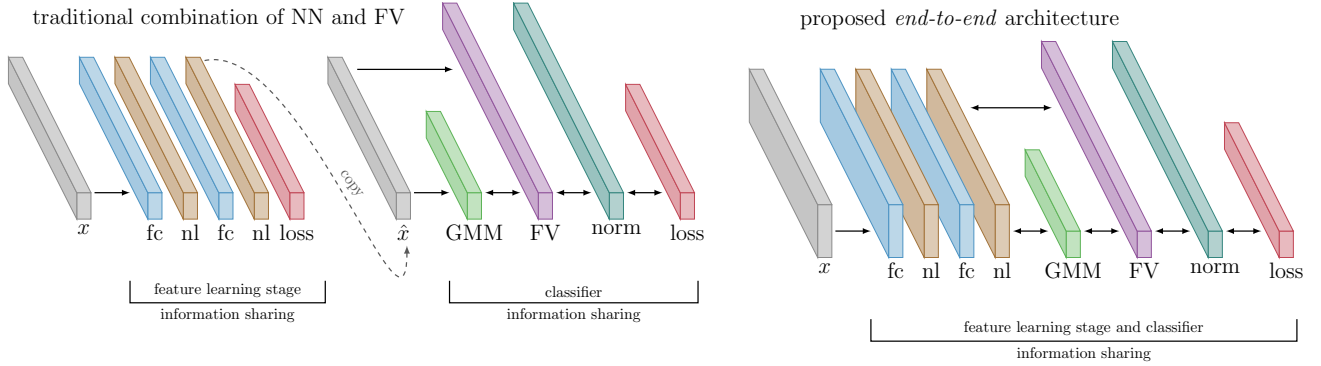

Figure 1: Difference between the traditional combination of deep neural networks (left) with fully-connected layers (fc), non-linearities (nl), normalization (norm) and FV compared to our proposed combination (right). Our approach fully integrates the GMM and FV layer enabling end-to-end with batched back-propagation.

$L_\lambda G_\lambda^X$, where $F_\lambda^{-1} = L_\lambda^T L_\lambda$ is the Cholesky decomposition of $F_\lambda^{-1}$. Although, a natural choice is $c_\lambda$ to be a Gaussian-Mixture-Model (GMM) as a weighted sum of $K$ Gaussians, an alternative is a hybrid Gaussian-Laplacian [10].

Assuming diagonal covariance matrices $\Sigma_k = \text{diag}(\sigma_k^2)$, $0 < \sigma_k^2 \in \mathbb{R}^D$ for efficient computations, learning a GMM with $K$ components gives $(2D+1)K$ tunable parameters

$$\lambda_1 \ldots, \lambda_K, \sigma_1, \ldots, \sigma_K \in \mathbb{R}^D, \mu_1, \ldots, \mu_K \in \mathbb{R}^D,$$

where $\lambda_k$ denotes the prior probability of Gaussian $\mathcal{N}(\cdot; \mu_k, \Sigma_k)$. For practical purposes all necessary computations are done in the log-domain using the logarithm of the multivariate normal distribution. For a fixed $x_t$ we abbreviate $c_j := \mathcal{N}(x_t; \mu_j, \sigma_j^2)$. The soft assignment or *posterior probability* for arbitrary but fixed $x_t \in X$ and a GMM with $K$ components is given by

$$\gamma_k(x_t) = \frac{\lambda_k \, \mathcal{N}(x_t; \mu_k, \sigma_k^2)}{\sum_{j=1}^K \lambda_j \, \mathcal{N}(x_t; \mu_j, \sigma_j^2)} \in [0, 1],$$

which can be computed using "logsumexp-trick" to avoid explicit computation of $c_j$. Computing

$$\mathcal{F}_{\lambda_k}^X = \frac{1}{T\sqrt{\lambda_k}} \sum_{t=1}^T (\gamma_k(x_t) - \lambda_k) \qquad (2)$$

$$\mathcal{F}_{\mu_k}^X = \frac{1}{T\sqrt{\lambda_k}} \sum_{t=1}^T \left( \gamma_k(x_t) \left( \frac{x_t - \mu_k}{\sigma_k} \right) \right) \qquad (3)$$

$$\mathcal{F}_{\sigma_k^2}^X = \frac{1}{T\sqrt{\lambda_k}} \sum_{t=1}^T \left( \gamma_k(x_t) \frac{1}{\sqrt{2}} \left( \frac{(x_t - \mu_k)^2}{\sigma_k^2} - 1 \right) \right) \qquad (4)$$

and concatenating these $K$ results $\mathcal{F}_{\lambda_k}^X \in \mathbb{R}, \mathcal{F}_{\mu_k}^X, \mathcal{F}_{\sigma_k^2}^X \in \mathbb{R}^D$ into a $(2D+1)K$ dimensional vector leads to the

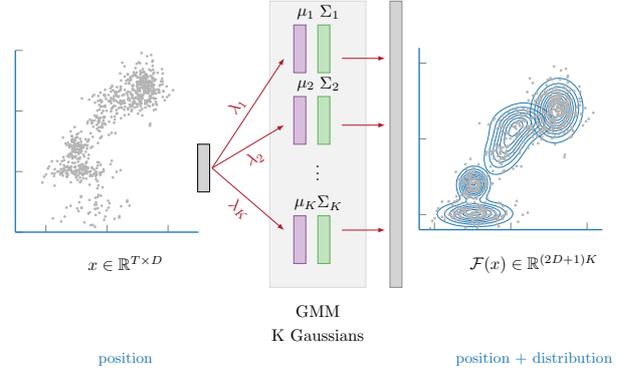

Figure 2: The FV encodes the data distribution fitted to a GMM in addition to the position of the feature.

Fisher-Vector representation ([15, 17]):

$$\mathcal{F} := \left( \mathcal{F}_{\lambda_1}^X, \ldots, \mathcal{F}_{\lambda_K}^X, \mathcal{F}_{\mu_1}^X, \ldots, \mathcal{F}_{\mu_k}^X, \mathcal{F}_{\sigma_1^2}^X, \ldots, \mathcal{F}_{\sigma_k^2}^X \right),$$

which is usually classified by a shallow architecture, *e.g.* a linear SVM. All calculus of vectors should be understood component-wise. This scheme is illustrated in Figure 2. Note, that $\mathcal{F}$ has a fixed dimension independent of $T$. As described in [22] eq. (2), (3) and (4) can be computed efficiently using pre-computed terms $S_k^p = \sum_t \gamma_k(x_t) x_t^p$ for $x_t \in \mathcal{X}, p = 0, 1, 2$. In our prototype implementation we were able to exploit the massively parallel nature of this step, which consists of several reductions.

Finally, a function composition of *power-normalization*

$$x \mapsto \text{sign}(x) \, |x|^\alpha, \quad \alpha \in (0, 1] \qquad (5)$$

and *L2-normalization*

$$x \mapsto x \, \|x\|_2^{-1} \qquad (6)$$

applied to $\mathcal{F}$ improves the performance [17] of the SVM+FV combination.

**Backpropagation** A neural network is usually a composition of functions $F := F_n \circ F_{n-1} \circ \cdots \circ F_1$ represented as stacked layers:

$$F_j : \mathbb{R}^{d_1} \to \mathbb{R}^{d_2}, \quad X_{j-1} \mapsto X_j := F_j(W_j, X_{j-1}),$$

where $X_{j-1} \in \mathbb{R}^{d_1}$ is the input to the $j$-th layer and $W_j$ is a collection of tunable parameters. Given $X_0$, the objective of the training phase for a neural network is to minimize a loss function $L(x, F(x))$ wrt. $W_{j=1,\dots,n}$ most often using stochastic gradient descent optimization.

Given partial derivatives wrt. $X_j$ it is possible to compute the partial derivatives wrt. $X_{j-1}, W_j$ using the chain rule

$$\frac{\partial F_j}{\partial W_j} = \frac{\partial}{\partial W} F_j(W_j, X_{j-1}) \frac{\partial F_{j+1}}{\partial X_j} \qquad (7)$$

$$\frac{\partial F_j}{\partial X_{j-1}} = \frac{\partial}{\partial X} F_j(W_j, X_{j-1}) \frac{\partial F_{j+1}}{\partial X_j} \qquad (8)$$

as described in [18].

## 4. Method

In the following, we describe all necessary steps for the Fisher-Vector layer to integrate this layer into back-propagation training.

**Fisher-Vector update rules:** Since equations (2)-(4) are differentiable in $(\lambda_j, \mu_j, \sigma_j^2)_{j=1,\dots,K}$, these parameters can be optimized using the gradients $\frac{\partial}{\partial x} \ln c_j, \frac{\partial}{\partial \mu} \ln c_j, \frac{\partial}{\partial \sigma^2} \ln c_j$ and the derivatives of FV wrt. $\ln c_j$ by exploiting the chain-rule.

Elementary calculation yields all derivatives of (2), (3), (4) wrt. to the GMM parameters and $x$, namely[1]

$\frac{\partial \mathcal{F}^X_{\lambda_k}}{\partial \lambda_s}, \frac{\partial \mathcal{F}^X_{\mu_k}}{\partial \lambda_s}, \frac{\partial \mathcal{F}^X_{\sigma_s^2}}{\partial \lambda_k}, \frac{\partial [\mathcal{F}^X_{\mu_k}]_d}{\partial [\mu_s]_e}, \frac{\partial [\mathcal{F}^X_{\sigma_k^2}]_d}{\partial [\mu_s]_e}, \frac{\partial [\mathcal{F}^X_{\mu_k}]_d}{\partial [\sigma_s^2]_e}, \frac{\partial [\mathcal{F}^X_{\sigma_k}]_d}{\partial [\sigma_s^2]_e},$
$\frac{\partial \mathcal{F}^X_{\lambda_k}}{\partial [x]_e}, \frac{\partial [\mathcal{F}^X_{\mu_k}]_d}{\partial [x]_e}, \frac{\partial [\mathcal{F}^X_{\mu_k}]_d}{\partial [x]_e}.$

See the appendix for the formulas. We use gradient descent to update parameters in the backward step.

One part of the gradient for the feature learning stage is given by

$$\frac{\partial}{\partial [x]_e} \left[\mathcal{F}^X_{\mu_k}\right]_d = \frac{1}{\sqrt{\lambda_k}} \left( \frac{\partial \gamma_k(x_t)}{\partial [x]_e} [\alpha_k(x_t)]_d + \delta_{ed} \left[\frac{\gamma_k(x)}{\sigma_k}\right]_d \right),$$

---
[1]Consult our prototype implementation of symbolic derivatives in the supplemental material.

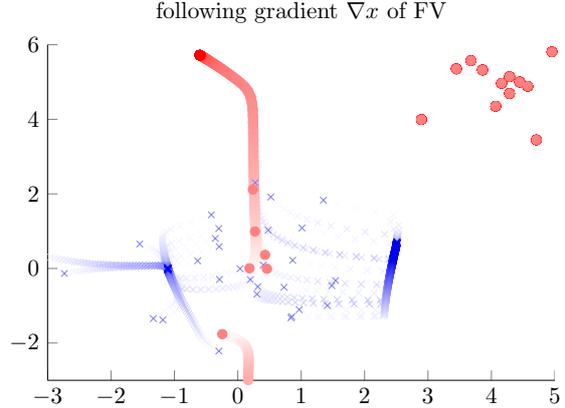

Figure 3: Although, it is not possible to separate both classes using the Fisher-Kernel, learning a transformation of $x$ facilitates the classification of points. Here, each data point $x$ is shifted along the FV-gradient (darker means later in the training process). After the optimization the points are clearly separable.

where

$$\frac{\partial}{\partial x_i} \gamma_k = \gamma_k(x_i) \left( -\beta_k(x_i) + \sum_{n=1}^{K} \beta_n(x_i)\gamma_n(x_i) \right), \quad (9)$$

$$\beta_k(x_i) := \frac{x_i - \mu_k}{\sigma_k^2} \qquad (10)$$

and Kronecker-$\delta_{ab} := (a{=}{=}b)$. Therefore, any update will push descriptor $x_i$ into the direction of the $k$-th Gaussian weighted by the posterior $\gamma_k(x_i)$ and $\frac{1}{\sigma_k^2}$. The effect of this gradient $\frac{\partial}{\partial [x]_e} \left[\mathcal{F}^X_{\mu_k}\right]_d$ is illustrated in Figure 3 for a 2D dataset. There, we shifted each input following the FV-gradient. Notice, that Eq. (9) has a connection to *inverse-variance weighting*.

**Satisfying the GMM parameter constraints:** It is crucial to not violate the GMM parameter assumptions $\sigma_k^2, \lambda_k > 0, \sum_k \lambda_k = 1$ when applying the steepest descent rule. The authors [21] proposed to re-normalize the weights $\lambda_{j=1,\dots,K} = \overline{\lambda_j}(\sum_k \overline{\lambda_k})^{-1}$ in each iteration to satisfy constraints for $\lambda_j$. Instead, we internally model each weight $\lambda_k$ as a sum of sigmoid-functions:

$$\lambda_j(\nu_j) = \frac{[1 + \exp(-\nu_j)]^{-1}}{\sum_\ell [1 + \exp(-\nu_\ell)]^{-1}} \in (0, 1) \qquad (11)$$

and represent $\sigma_k^2$ as

$$\sigma_k^2(\zeta_j) = \varepsilon + \exp(\zeta_j) > 0 \qquad (12)$$

for some $\nu_j, \zeta_j \in \mathbb{R}$ and $\varepsilon > 0$, which allows to optimize objective (1) in an unconstrained setting in $\lambda, \mu, \sigma^2, x$ and provides a natural way to satisfy all GMM parameter constraints $\sigma_k^2 > \varepsilon, \lambda_j \in [0,1], \sum_\ell \lambda_\ell = 1$, without numerical issues or projection steps during gradient descent.

**Design of SVM gradient information:** Let $(x_i, y_i)_{i=1,...,N}$ some training data consisting of feature $x_i \in \mathbb{R}^D$ with label $y_i \in \{-1, +1\}$. The back-propagation process starts from the SVM layer, the standard C-SVM:

$$\min_{\theta,b} \frac{1}{\|\theta\|_2^2} + \frac{C}{N}\sum_{i=1}^N \max\{0,\, 1 - y_i(\langle\theta, x_i\rangle + b)\} \quad (13)$$

with regularization constant $1/C$ and hinge loss. The update information from (13) is sparse and only forces changes after a long training period, whenever the rare event occurs that a batch contains a miss-classified point. hence, using the non-differential hinge loss for back-propagation one ends up with a sub-gradient-method. Although, the quadratic hinge loss as in [21] is smoother, it also creates sparse gradients.

Switching to the quadratic loss $(1 - y_i(\langle\theta, x_i\rangle + b))^2$ for back-propagating a dense gradient information would move the data points to the margin (see Figure 4), which is unsuitable, since these clusters cannot be represented by Gaussian with diagonal covariance matrices. We use $-y\theta^T$ to shift all data towards the correct "side" away from the decision boundary – detached from the actual SVM formulation for classifying these points.

**Normalization layer:** The post-processing of FV, a function composition of (5) and (6), which refer to *normalization layer* can be expressed as

$$\phi(x) = \frac{\text{sign}(x)|x|^\alpha}{\|\text{sign}(x)|x|^\alpha\|_2}. \quad (14)$$

Its derivative $\frac{\partial F_j}{\partial X_{j-1}} = \phi'(X_{j-1})\frac{\partial F_{j+1}}{\partial X_j}$ for $\alpha = 0.5$ is given as

$$\phi'(x) = \frac{2}{\|x\|_1}(\mathbb{1}_D - \frac{\hat{x}\hat{x}^T}{\hat{x}^T\hat{x}})\mathbb{1}_{x\neq 0} \quad (15)$$

for $\hat{x} = \text{sign}(x)\sqrt{|x|}$ and identity matrix $\mathbb{1}_D$.

**Initialization:** Learning the GMM is done by k-means initialization until the convergence of the log-likelihood. Initializing the feature learning stage is the tricky part. When using a random transformation $W$ of the PCA-projected SIFT vectors $x$, the algorithm suffers from a bad initialization. Therefore, we revert these transformations to have exactly the same starting input for the Fisher layer by using features $\tilde{x}$ satisfying $x = \tanh(W\tilde{x} + b) \in [-1, +1]$ in our method. The parameter $W$ is initialized using *Xavier* [7] initialization.

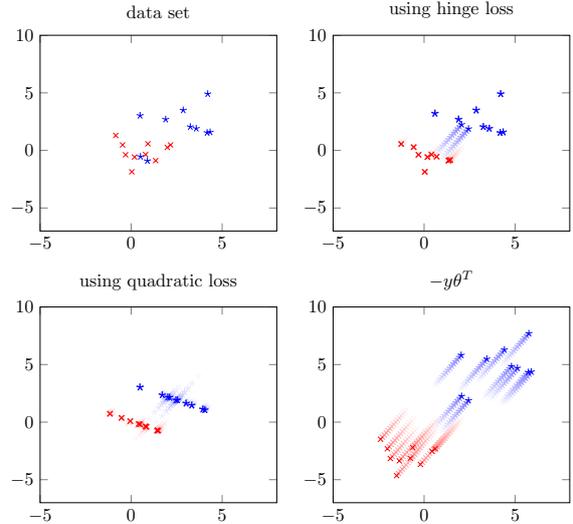

Figure 4: The update-effect of different loss (sub-) gradients in the SVM layer: The hinge loss produces sparse gradients. On the other hand a GMM with coordinates-independence assumption is not capable of representing the result of the quadratic loss. We propose to directly use $-y\theta^T$, which simply shifts each point in the correct direction away from the classification boundary.

## 5. Implementation and Evaluation

We now describe our prototype CPU/GPU-based implementation, the used datasets and the experimental evaluation of our approach. We evaluate the effectiveness of training additional parameters within the parameter $W$ in our experiments.

### 5.1. Real-World Dataset and Feature extraction

The publicly available Pascal Voc 2007 (Voc07) dataset [6] comprises 20 classes of objects to be recognized, split up into 5011 images for training and 4952 images for testing for each class.

As the initial fixed features representation we use dense SIFT features extracted at multiple scales from OpenCV normalized to $[-1, 1]$. Each of these 128 dimensional local features is projected onto the first 64 principal components similar to [21], [17]. We observed no decrease of performance when representing each image by randomly selected $10^4$ features per image instead of approx. $9 \cdot 10^4$ similar to [21], which reduces the storage costs from 204GB to 25GB. Hence, in our approach each back-propagation needs to compute more than 25GB derivatives per epoch.

### 5.2. Implementation Details

The implemented architecture consists of multiple layers, which process the input features in batches of 24 images for fast enough updates. Each image described by $T$ PCA-transformed 64 dimensional features is used to compute a 4128 dimensional Fisher-Vector, which was normalized as described in section 3. The underlying GMM contains 32 Gaussians. We trained the SVM (13) using stochastic dual coordinate ascent (SDCA) with $C := N$, which gives superior performance than applying PEGASOS [19] in the primal.

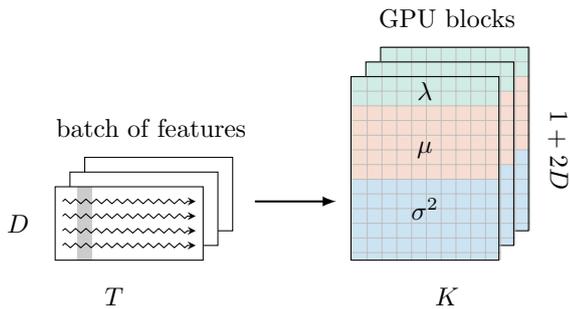

Figure 5: The feature set of each image is distributed to one GPU thread block. The GPU block computes the respective calculus for each feature in the set one after another.

| task | timing | | speedup of GPU |
|---|---|---|---|
| | Matlab | GPU | |
| $\mathcal{F}^X$ | 9.06s | 20ms | $\times 453$ |
| $\frac{\partial}{\partial \lambda, \mu, \sigma^2} \mathcal{F}^X$ | 18.9h | 10.79s | $\times 6306$ |
| $\frac{\partial}{\partial x} \mathcal{F}^X$ | 1.9h | 2.89s | $\times 2367$ |
| sum | 19.1h | 13.71s | $\times 5015$ |

Table 1: Timing comparison of vectorized MATLAB version and GPU implementation for a single batch of 24 images.

All remaining parameter updates are done by SGD with learning rate $\eta = 10^{-4}$. In contrast, [21] determines an optimal step size in each update.

### 5.3. GPU Implementation

The main challenge of a batch-wise end-to-end Fisher-Vector implementation is the highly non-linear mapping of each FV-batch back to feature and GMM space. Furthermore, the computation of those derivatives has a high complexity of $\mathcal{O}(K^2 D^2 T)$. Thus, a fast GPU implementation is indispensable to train the classifier in a feasible amount of time.

Our approach is well suited to take advantage of GPU parallism. There are two levels of granularity in our implementation. The first is by processing the images in parallel. In this fashion, each set of features is assigned to one block of threads. The second is by utilizing a block layout, which is shaped like the derivatives themselves. Hence, it is possible to process all elements of a feature in parallel while sweeping over the complete set of features. This procedure ensures that global memory access is kept at a minimum, while all computations are performed with on-chip memory. Figure 5 shows a scheme of this approach. For the implementation we used CUDA and a setting of four NVIDIA Titan X GPUs. The limiting factors in our GPU approach are the number of available registers and the size of shared memory.

**Timings** For computing the FV and respective derivatives, a comparison between our MATLAB and GPU implementation is shown in Table 1. The highly vectorized MATLAB code is running on a i5-2500 CPU with 3.30GHz. In this experiment, the performance is measured on one CPU core compared to one GPU. The batch size is 24 with $10^4$ features each. The reported GPU timings also include all necessary memory transfers between host and device system. Furthermore, both variants compute dense vectors and have no criteria to omit data.

We allow for concurrent Kernel execution, which adapts better for the available resources. The performance for different batch sizes is illustrated in Figure 6, where we use all four GPUs as well as all four cores of the i5 CPU.

In summary, the MATLAB implementation with four cores takes more than 4.7 hours to compute a complete forward and backward step of the Fisher layer for batches of size 24. Our multi-GPU implementation reduces the amount of computation time to less than 5.2 seconds. Note, that for the PASCAL VOC 2007 challenge a complete sweep over the training data comprises 208 batches of 24 images, which still results in about 18 minutes runtime per epoch.

### 5.4. Experimental Results

To ensure fair evaluation we tested our methods against the baseline method of only training the SVM classifier. To compare our method against [21], we re-implemented their algorithm up to batch-wise updates. As an evaluation metric, we use the *average precision* (AP), corresponding to the area under the precision-recall curve.

In detail, our training process consists of different phases. First, the initial training (with at most 15 epochs, $\approx 3000$ iterations) of SVM was done. We observed the convergence of the SVM within this phase for all 20 experiments, i.e., the duality gap is less than $0.01$. This initial training guarantees already good performance as reported in the first column of Table 2, ie. only $\theta$ was learned. Note, that our initial training already significantly outperforms the reported base-line results from [21].

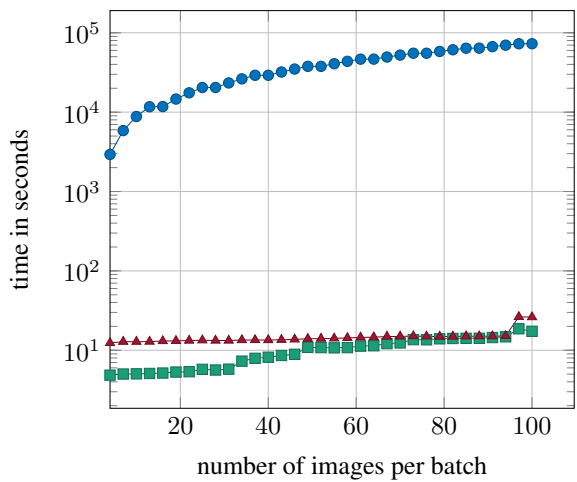

Figure 6: Timing for CPU(—●—), async. GPU (—■—) and sync. GPU (—▲—) for a complete forward and backward batch-computation of the Fisher layer. Smaller is better. Hence, all four GPUs are saturated for 24 images per batch.

| class | AP (when optimizing parameters) | | |
|---|---|---|---|
| | $\theta$ | $\theta, \lambda_k, \mu_k, \sigma_k^2$ | $\theta, \lambda_k, \mu_k, \sigma_k^2, (W,b)$ |
| aeroplane | 74.1 | 75.2 (+1.1) | **77.3 (+3.2)** |
| bicycle | 55.1 | 55.7 (+0.6) | **59.5 (+4.4)** |
| bird | 39.6 | 40.8 (+1.4) | **41.9 (+2.3)** |
| boat | 68.2 | 68.5 (+0.3) | **69.8 (+1.6)** |
| bottle | 24.2 | 24.8 (+0.6) | **25.0 (+0.8)** |
| bus | 56.4 | 56.9 (+0.5) | **59.0 (+2.6)** |
| car | 74.5 | 74.9 (+0.4) | **77.8 (+3.3)** |
| cat | 50.2 | 50.8 (+0.6) | **53.8 (+3.6)** |
| chair | 49.9 | 50.6 (+0.7) | **51.8 (+1.9)** |
| cow | 30.1 | 32.2 (+1.1) | **34.3 (+4.2)** |
| dining table | 42.1 | 43.1 (+1.0) | **44.5 (+2.4)** |
| dog | 34.3 | 35.0 (+0.7) | **38.2 (+3.9)** |
| horse | 75.3 | 75.2 (-0.1) | **76.8 (+1.5)** |
| motorbike | 55.1 | 55.3 (+0.2) | **57.9 (+2.8)** |
| person | 81.1 | 81.3 (+0.2) | **82.6 (+1.5)** |
| pottedplant | 23.6 | 24.7 (+1.4) | **27.5 (+3.9)** |
| sheep | 37.7 | 38.9 (+1.2) | 38.9 (+1.2) |
| sofa | 48.9 | 49.8 (+0.9) | **51.1 (+1.2)** |
| train | 75.6 | 75.9 (+0.3) | **77.8 (+2.2)** |
| tvmonitor | 46.8 | 46.7 (-0.1) | **49.5 (+2.7)** |
| mAP | 52.1 | 52.8 (+0.7) | **54.7 (+2.6)** |

Table 2: Results on the PascalVoc2007-Database, when optimizing different parameter sets.

From this, we start to evaluate further training of GMM parameters $(\lambda_k, \mu_k, \sigma_k^2)_{k=1,...,K}$ and $\theta$. Training the Fisher kernel alone slightly improves previous performance, at the price of much higher computation time. The corresponding mean AP is comparable to [21]. Starting again from the strong initial training, we now trained all parameters from the GMM and SVM including the feature mapping of our linear layer with Tanh activation. Allowing the first layer to update parameters $W$ and $b$ makes the solution process more flexible allowing the fitting of $\tanh(Wx+b)$ better to the GMM parameters.

Our approach of optimizing all parameters increases the gain of training only GMM parameters by more than three times from 52.8 (+0.7) to 54.7 (+2.6). Remarkably, the total computational effort increases only by factor 1.2, compared to the methods of [21].

## 6. Conlusion and Outlook

We introduce feature learning in combination with Fisher-Vectors in the fashion of neural networks, which paves the way for a wider range of applications for Fisher-Vectors. Our interpretation of the Fisher-Kernel as a module with a forward and backward pass allows end-to-end training-architectures to benefit from this data distribution encoding scheme. Analogously to the huge impact of GPU implementations in deep learning methods, we expect further progress for GPU-accelerated FV implementations in combinations with other methods.

We believe that this approach enables several future directions. One interesting idea might be the embedding of Fisher-Vector modules in deep neural networks at arbitrary positions. Extracted features from convolution filters can be pooled applying the Fisher layer instead of a fully connected layer. We are planning to add our implementation to popular deep learning frameworks like Caffe or mxnet. Another interesting way of using Fisher-Vectors, is to combine our approach with the work of [20] to train stacked Fisher layers in deep architecture.

Currently, the success of FV depends on robust downsampling approaches like PCA. A batch-wise replacement would facilitate a large-scale end-to-end pipeline including Fisher-Vector training. Solving challenges like pre-training a Gaussian mixture model in a batch-wise online mode would help to realize a fully embedded Fisher layout in a classical back-propagation training.

## 7. Appendix

The derivatives of the Fisher-Vector layer are given by:

$$\frac{\partial}{\partial \lambda_s} \mathcal{F}^X_{\lambda_k} = \sum_{t=1}^T \frac{\delta_{sk}(\gamma_k - \lambda_k) - 2\gamma_s \gamma_k}{2\lambda_s \sqrt{\lambda_k}}$$

$$\frac{\partial}{\partial \lambda_s} \mathcal{F}^X_{\mu_k} = \sum_{t=1}^T \frac{\gamma_k \alpha_k (\delta_{sk} - 2\gamma_s)}{2\lambda_s \sqrt{\lambda_k}}$$

$$\frac{\partial}{\partial \lambda_k} \mathcal{F}^X_{\sigma_s^2} = \sum_{t=1}^T \frac{\gamma_k (\alpha_k^2 - 1)(\delta_{ks} - 2\gamma_s)}{2\lambda_s \sqrt{2\lambda_k}}$$

$$\frac{\partial}{\partial [\mu_s]_e} \mathcal{F}^X_{\lambda_k} = \sum_{t=1}^T \frac{\gamma_k (\delta_{ks} - \gamma_s) \alpha_s}{[\sigma_s]_e \sqrt{\lambda_k}}$$

$$\frac{\partial}{\partial [\mu_s]_e}\left[\mathcal{F}^X_{\mu_k}\right]_d = \sum_{t=1}^{T} \frac{\gamma_k\left((\delta_{ks}-\gamma_s)[\alpha_s]_e[\alpha_k]_d - \delta_{sk}\delta_{ed}\right)}{[\sigma_s]_e\sqrt{\lambda_k}}$$

$$\frac{\partial}{\partial [\mu_s]_e}\left[\mathcal{F}^X_{\sigma_k^2}\right]_d = \sum_{t=1}^{T} \frac{\gamma_k\left([\alpha_s]_e\left((\delta_{ks}-\gamma_s)([\alpha_k^2]_d - 1) - 2\delta_{de}\delta_{sk}\right)\right)}{[\sigma_s]_e\sqrt{2\lambda_k}}$$

$$\frac{\partial}{\partial [\sigma_s^2]_e}\mathcal{F}^X_{\lambda_k} = \sum_{t=1}^{T} \frac{\gamma_k(\delta_{ks}-\gamma_s)\left([\alpha_s^2]_e - 1\right)}{2[\sigma_s^2]_e\sqrt{\lambda_k}}$$

$$\frac{\partial}{\partial [\sigma_s^2]_e}\left[\mathcal{F}^X_{\mu_k}\right]_d = \sum_{t=1}^{T} \frac{\gamma_k[\alpha_k]_d\left((\delta_{ks}-\gamma_s)[\alpha_s^2-1]_e - \delta_{sk}\delta_{ed}\right)}{2[\sigma_s^2]_e\sqrt{\lambda_k}}$$

$$\frac{\partial}{\partial [\sigma_s^2]_e}\left[\mathcal{F}^X_{\sigma_k}\right]_d = \sum_{t=1}^{T} \frac{\gamma_k\left((\delta_{ks}-\gamma_s)[\alpha_s^2-1]_e \cdot [\alpha_k^2-1]_d - 2\delta_{ks}\delta_{ed}[\alpha_k]_d^2\right)}{2[\sigma_s^2]_e\sqrt{2\lambda_k}}$$

$$\frac{\partial}{\partial [x]_e}\mathcal{F}^X_{\lambda_k} = \frac{1}{\sqrt{\lambda_k}}\frac{\partial \gamma_k(x)}{\partial [x]_e}$$

$$\frac{\partial}{\partial [x]_e}\left[\mathcal{F}^X_{\mu_k}\right]_d = \frac{1}{\sqrt{\lambda_k}}\left(\frac{\partial \gamma_k}{\partial [x]_e}[\alpha_k]_d + \delta_{ed}\left[\frac{\gamma_k(x)}{\sigma_k}\right]_d\right)$$

$$\frac{\partial}{\partial [x]_e}\left[\mathcal{F}^X_{\sigma_k^2}\right]_d = \frac{1}{\sqrt{2\lambda_k}}\left(\frac{\partial \gamma_k(x)}{\partial [x]_e}[\alpha_k^2 - 1]_d + 2\delta_{ed}\gamma_k(x)[\beta_k(x_t)]_d\right).$$

We abbreviate $\gamma_\ell(x_t), \alpha_\ell(x_t) = \frac{x_t-\mu_\ell}{\sigma_\ell}$ by dropping the argument. Note, most expressions are sums of dyadic products. Therefore, all gradients are build up on the gradients of the soft-assignment function $\gamma_k$ which can be pre-computed:

$$\frac{\partial}{\partial \lambda_s}\gamma_k = \gamma_k\left(\frac{\delta_{ks}}{\lambda_k} - \frac{\gamma_s}{\lambda_s}\right)$$

$$\frac{\partial}{\partial \mu_s}\gamma_k = \gamma_k(\delta_{ks}-\gamma_s)\left(\frac{x-\mu_s}{\sigma_s^2}\right)$$

$$\frac{\partial}{\partial \sigma_s^2}\gamma_k = \gamma_k(\delta_{ks}-\gamma_s)\left(\frac{(x-\mu_s)^2}{2\sigma_s^4} - \frac{1}{2\sigma_s^2}\right).$$

## References

[1] H. Bay, A. Ess, T. Tuytelaars, and L. Van Gool. Speeded-up robust features (surf). *Comput. Vis. Image Underst.*, 110(3):346–359, June 2008. 1

[2] S. Bell, P. Upchurch, N. Snavely, and K. Bala. Material recognition in the wild with the materials in context database. *CoRR*, abs/1412.0623, 2014. 1

[3] M. Cimpoi, S. Maji, and A. Vedaldi. Deep convolutional filter banks for texture recognition and segmentation. *CoRR*, abs/1411.6836, 2014. 1

[4] D. C. Ciresan, U. Meier, and J. Schmidhuber. Multi-column deep neural networks for image classification. *CoRR*, abs/1202.2745, 2012. 1

[5] M. Dixit, S. Chen, D. Gao, N. Rasiwasia, and N. Vasconcelos. Scene classification with semantic fisher vectors. June 2015. 1, 2

[6] M. Everingham, L. Gool, C. K. Williams, J. Winn, and A. Zisserman. The pascal visual object classes (voc) challenge. *Int. J. Comput. Vision*, 88(2):303–338, June 2010. 5

[7] X. Glorot and Y. Bengio. Understanding the difficulty of training deep feedforward neural networks. In *In Proceedings of the International Conference on Artificial Intelligence and Statistics (AISTATS10). Society for Artificial Intelligence and Statistics*, 2010. 5

[8] T. Jaakkola and D. Haussler. Exploiting generative models in discriminative classifiers. In *In Advances in Neural Information Processing Systems 11*, pages 487–493. MIT Press, 1998. 2

[9] H. Jegou, M. Douze, C. Schmid, and P. Perez. Aggregating local descriptors into a compact image representation. In *Computer Vision and Pattern Recognition (CVPR), 2010 IEEE Conference on*, pages 3304–3311, June 2010. 1

[10] B. Klein, G. Lev, G. Sadeh, and L. Wolf. Fisher vectors derived from hybrid gaussian-laplacian mixture models for image annotation. *CoRR*, abs/1411.7399, 2014. 3

[11] L. Liu, C. Shen, L. Wang, A. van den Hengel, and C. Wang. Encoding high dimensional local features by sparse coding based fisher vectors. In Z. Ghahramani, M. Welling, C. Cortes, N. Lawrence, and K. Weinberger, editors, *Advances in Neural Information Processing Systems 27*, pages 1143–1151. Curran Associates, Inc., 2014. 1, 2

[12] D. G. Lowe. Distinctive image features from scale-invariant keypoints. *Int. J. Comput. Vision*, 60(2):91–110, Nov. 2004. 1

[13] A. Oliva and A. Torralba. Modeling the shape of the scene: A holistic representation of the spatial envelope. *Int. J. Comput. Vision*, 42(3):145–175, May 2001. 1

[14] F. Perronnin and C. Dance. Fisher kernels on visual vocabularies for image categorization. In *Computer Vision and Pattern Recognition, 2007. CVPR '07. IEEE Conference on*, pages 1–8, June 2007. 1

[15] F. Perronnin and C. Dance. Fisher kernels on visual vocabularies for image categorization. In *Computer Vision and Pattern Recognition, 2007. CVPR '07. IEEE Conference on*, pages 1–8, June 2007. 3

[16] F. Perronnin and D. Larlus. Fisher vectors meet neural networks: A hybrid classification architecture. June 2015. 2

[17] F. Perronnin, J. Snchez, and T. Mensink. Improving the fisher kernel for large-scale image classification. In *Proceedings of the 11th European Conference on Computer Vision: Part IV*, ECCV'10, pages 143–156, Berlin, Heidelberg, 2010. Springer-Verlag. 3, 4, 5

[18] D. E. Rumelhart, G. E. Hinton, and R. J. Williams. Neurocomputing: Foundations of research. chapter Learning Representations by Back-propagating Errors, pages 696–699. MIT Press, Cambridge, MA, USA, 1988. 4

[19] S. Shalev-Shwartz, Y. Singer, and N. Srebro. Pegasos: Primal estimated sub-gradient solver for svm. In *Proceedings of the 24th International Conference on Machine Learning*, ICML '07, pages 807–814, New York, NY, USA, 2007. ACM. 6

[20] K. Simonyan, A. Vedaldi, and A. Zisserman. Deep fisher networks for large-scale image classification. In C. Burges, L. Bottou, M. Welling, Z. Ghahramani, and K. Weinberger, editors, *Advances in Neural Information Processing Systems 26*, pages 163–171. Curran Associates, Inc., 2013. 2, 7

[21] V. Sydorov, M. Sakurada, and C. H. Lampert. Deep fisher kernels - end to end learning of the fisher kernel gmm parameters. June 2014. 2, 4, 5, 6, 7

[22] J. Snchez, F. Perronnin, T. Mensink, and J. Verbeek. Image classification with the fisher vector: Theory and practice.

*International Journal of Computer Vision*, 105(3):222–245, 2013. 2, 3

[23] V. Vapnik and O. Chapelle. Bounds on error expectation for support vector machines. *Neural Comput.*, 12(9):2013–2036, Sept. 2000. 1, 2

[24] D. Yoo, S. Park, J. Lee, and I. Kweon. Fisher kernel for deep neural activations. *CoRR*, abs/1412.1628, 2014. 1